\documentclass{article}
\pdfoutput=1

\usepackage{ijcai24}

\usepackage{times}
\usepackage{soul}
\usepackage{url}
\usepackage[hidelinks]{hyperref}
\usepackage[utf8]{inputenc}
\usepackage[small]{caption}
\usepackage{graphicx}
\usepackage{amsmath}
\usepackage{amsthm}
\usepackage{booktabs}
\usepackage{algorithm}
\usepackage{algorithmic}
\usepackage[switch]{lineno}
\usepackage{amsfonts,amssymb}

\title{Self-supervised Pretraining for Decision Foundation Model:
Formulation, Pipeline and Challenges}

\author{
Xiaoqian Liu$^1$
\and
Jianbin Jiao$^1$
\and
Junge Zhang $^{1,2}$\footnote{\* Corresponding author: Junge Zhang}
\affiliations
$^1$University of Chinese Academy of Sciences\\
$^2$Institute of Automation, Chinese Academy of Sciences\\
\emails
liuxiaoqian23@mails.ucas.ac.cn,
jiaojb@ucas.ac.cn,
jgzhang@nlpr.ia.ac.cn
}

\begin{document}

\maketitle

\begin{abstract}
Decision-making is a dynamic process requiring perception, memory, and reasoning to make choices and find optimal policies. 
Traditional approaches to decision-making suffer from sample efficiency and generalization, while large-scale self-supervised pretraining has enabled fast adaptation with fine-tuning or few-shot learning in language and vision.
We thus argue to integrate knowledge acquired from generic large-scale self-supervised pretraining into downstream decision-making problems.
We propose Pretrain-Then-Adapt pipeline and survey recent work on data collection, pretraining objectives and adaptation strategies for decision-making pretraining and downstream inference.
Finally, we identify critical challenges and future directions for developing decision foundation model with the help of generic and flexible self-supervised pretraining.
   
\end{abstract}

\section{Introduction}
Human life is all about making decisions. It occurs in a wide variety of scenarios and intelligent agents have been developed to help human with real-world decision-making tasks, such as traffic control \cite{liang2019deep}, energy management \cite{nakabi2021deep}, and drug discovery \cite{zhou2019optimization}. Dominant paradigms to train agents include reinforcement learning (RL), imitation learning (IL) and planning. Recent advances in these algorithms have achieved superhuman performance in mastering the game of Go \cite{silver2017mastering}, playing Atari video games directly from pixels \cite{schwarzer2023bigger}, and autonomous control of robotic locomotion and manipulation from sensory data \cite{brohan2022rt}. However, traditional approaches to agent training suffer from sample efficiency and generalization as numerous interactions with the environment are demanded and the agent is task- and domain-specific. In this paper, we argue that self-supervised pretraining with downstream adaptation is one way to alleviate the issues. We survey recent work especially on multi-task offline pretraining via Transformer \cite{vaswani2017attention} and propose future research directions towards general large-scale pretraining models for decision-making.

Self-supervised pretraining has enabled large sequence models to realize few-shot or even zero-shot adaptation in natural language processing (NLP) \cite{Achiam2023GPT4TR} and computer vision (CV) tasks \cite{bai2023sequential}. Through pretraining on large generic corpora or visual data (images and videos), knowledge about the world and human society is learned which can be utilized in various downstream task learning with few samples so as to improve sample efficiency and generalization. In NLP, self-supervised sequential modeling using the objective of next word prediction or random masking has produced powerful pretrained models for text generation \cite{brown2020language} or context understanding \cite{devlin2018bert}. In CV, unsupervised contrastive learning \cite{chen2020simple,he2020momentum} has shown effectiveness in data augmentation to improve visual representation learning. Recently, supervised pretraining and online pretraining have made some progress in decision making by leveraging large-scale expert demonstrations to perform imitation learning \cite{reed2022generalist,brohan2022rt,lee2023supervised} or involving self-supervised prediction or skill discovery in online exploration \cite{pathak2017curiosity,eysenbach2018diversity}.
However, large-scale self-supervised pretraining in multi-task offline settings remains a significant research challenge. 

Recent advances in self-supervised (unsupervised) pretraining for decision-making allow the agent to learn without reward signals and in accordance with offline RL \cite{levine2020offline}, encourage the agent to learn from sub-optimal offline data. However, previous attempts at self-supervised RL pretraining have mostly been limited to a single task \cite{yang2021representation}, or performing pretraining and fine-tuning within the same task \cite{schwarzer2021pretraining}, which is not generic and flexible for adapting to varieties of downstream tasks. Therefore, in this paper, we focus on multi-task offline pretraining for decision foundation model where the pretraining data is task-irrelevant and collected from diverse environments, and the downstream adaptation concerns different tasks with varying dynamics or reward functions and domains with distinct state or action spaces. In addition, we highlight self-supervised representation learning in pretraining for decision-making to probe what knowledge can be learned and whether it can be transferred to solve decision-making tasks, despite that fully unsupervised RL training requires learning self-supervised representation as well as general policies via a single model \cite{laskin2021urlb,reed2022generalist,brohan2022rt}.

We first provide background on sequential decision-making and techniques for sequential modeling and self-supervised learning for RL (\S2.1). We then propose Pretrain-Then-Adapt pipeline to formulate the problem of self-supervised pretraining for decision foundation model (\S2.2). Based on this pipeline, we review recent work considering three steps: data collection (\S3), self-supervised pretraining (\S4), and downstream adaptation (\S5). For data collection, we summarize environments and tasks for decision-making pretraining and differences between pretraining dataset and downstream data. For self-supervised pretraining, we introduce tokenization strategies and pretraining objectives specifically designed for decision-making tasks used in recent work. For downstream adaptation, we categorize inference tasks to evaluate pretraining performance and introduce two strategies of adapting the pretrained model to downstream inference task including fine-tuning and zero-shot generalization. 
We conclude by identifying key challenges for general self-supervised pretraining for decision foundation model (\S6) and advocate new research methods and evaluation framework for developing intelligent agents utilizing knowledge acquired from the foundation model.

\begin{figure*}[h]
    \centering
    \includegraphics[scale=0.53]{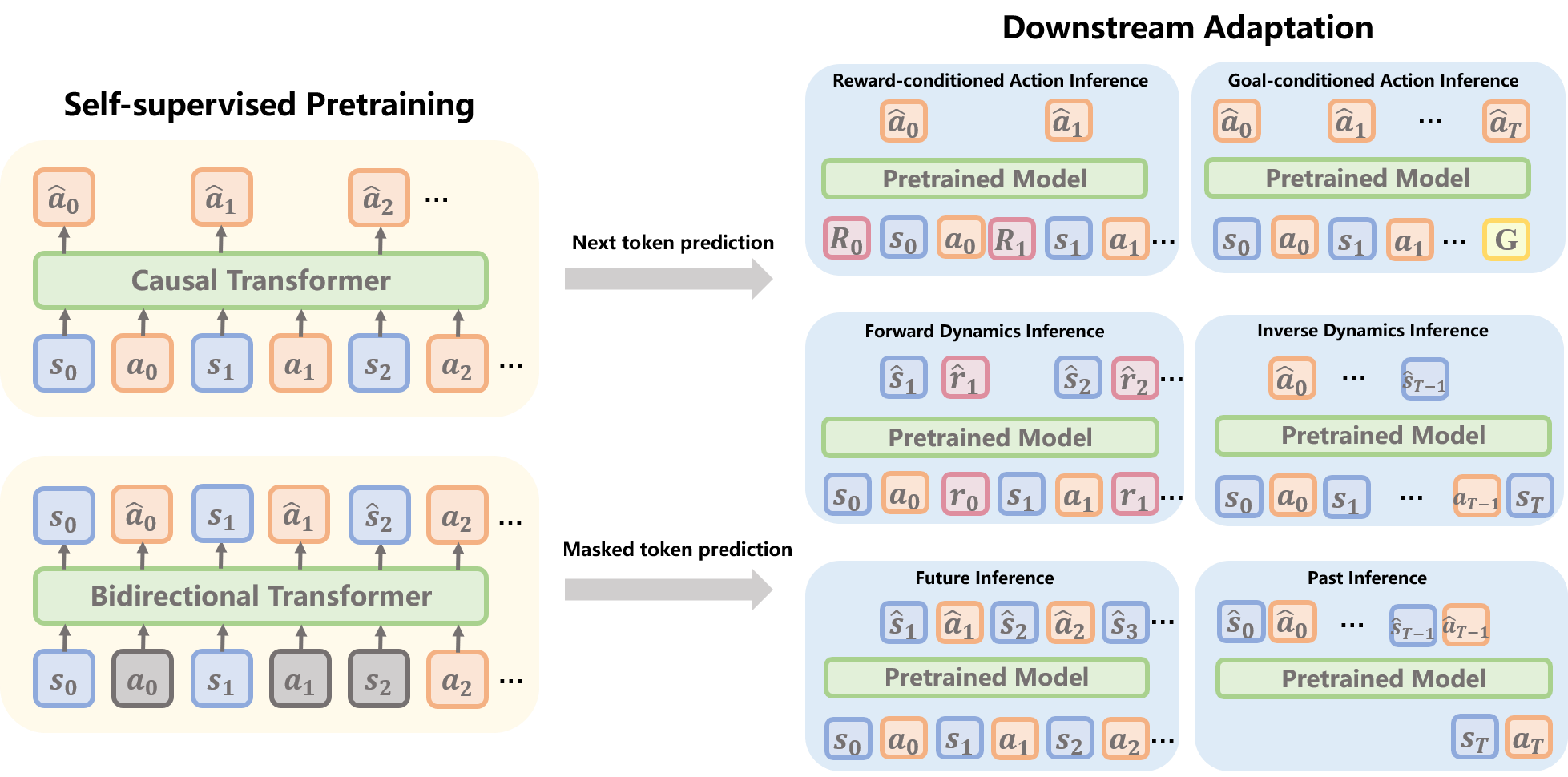}
    \caption{Pretrain-then-Adapt pipeline for decision foundation model. \textbf{Left}: Self-supervised pretraining involves two basic pretraining objectives based on Transformer architecture: next token prediction and masked token prediction. However, using different RL components, various pretraining objectives can be proposed for pretraining decision foundation models as introduced in section \ref{sec:objective}. \textbf{Right}: Downstream inference tasks can be roughly divided into three categories: action inference, dynamcis inference and trajectory inference. Details about each inference task can be referred to section \ref{sec:infer task}. Different colors denote different data modality in trajectory sequences. The grey color means the token is masked.}
    \label{fig:pipeline}
\end{figure*}

\section{Problem Formulation}

\subsection{Preliminaries}

\subsubsection{Sequential Decision-Making}
Sequential decision-making refers to the process of making a series of decisions to achieve a goal in continuous time, based on previous actions and observations while considering possible future states and rewards. The process can be formulated as a markov decision process (MDP) $M=<S,A,T,R,\gamma>$,
where $S$ is the state space, $A$ the action space, $T$ the dynamics
transition function $T:S\times A\times S\to [0,1)$, $R$ the reward
function $R:S\times A\times S\to \mathbb{R}$, and $\gamma \in [0, 1)$ is a discount
factor for calculating cumulative rewards. Sometimes the underlying state is not accessible to the agent (e.g, an image-based Atari game), and the process can be modified as a partially observable MDP $M=<S,A,T,R,\mathcal{O},E>$, where $\mathcal{O}$
is the observation space, and $E(o|s)$ denotes the observation emission function. Solutions to sequential decision-making often involve RL \cite{10.5555/3312046}, which aims to learn an optimal policy $\pi(a|s)$ that maximizes the expected discounted cumulative rewards $R=\sum_{k=0}^{\infty}\gamma^kr_{k+1}$.

\subsubsection{RL via Sequence Modeling}
Conventional RL focuses on online interactions with an environment, where an agent learns to maximize cumulative rewards through trial and error. The agent is able to get instant feedback from the environment but it is sample-inefficient especially in real-world applications. Offline RL can mitigate the issue by reusing past experiences from a static dataset. However, due to error propagation and value overestimation, it is challenging to train an offline RL agent via dynamic programming. By conditioning an autoregressive model on previous trajectories containing past states, actions and target return tokens, Decision Transformer \cite{chen2021decision} formulates RL as a sequence modeling problem. Such a formulation draws upon the simplicity and scalability of Transformer without the need to fit value functions or compute policy gradients.

\subsubsection{Self-supervised Learning for RL}
For RL, sample efficiency and generalization are two key properties and can be realized and improved by sufficient representation learning. Similar to pretraining in language \cite{devlin2018bert} and vision \cite{bao2021beit}, pretraining on a diverse, large amount of online or offline experiences benefits downstream task adaptation with pretrained representation that can be generalized across domains and tasks. Previous pretraining for RL mostly relies on expert demonstrations \cite{stooke2021decoupling} for one single task \cite{schwarzer2021pretraining}, yet in real-world scenarios, large amount of unlabeled and sub-optimal data (without task-specific reward) can be collected from multiple environments and tasks. Self-supervised learning is an effective and task-agnostic method to learn representation from numerous unlabeled data as shown in large language or vision model pretraining \cite{brown2020language,bai2023sequential}.

Self-supervised learning for RL typically involves representation learning of separate RL components, such as state representation \cite{yarats2021image}, action representation \cite{chandak2019learning}, reward representation \cite{ma2022vip}, policy representation \cite{tang2022inputting} and environment or task representation \cite{wang2023meta}. However, based on RL via sequence modeling, self-supervised learning for RL can also be formulated as a unified sequence modeling of various RL components, thus simultaneously learn representation and extract knowledge concerning different aspects of sequential decision-making processes, including temporal, causal and dynamics information.

\subsection{Pretrain-Then-Adapt}
In this survey, we focus on the pretrain-then-adapt pipeline for decision foundation model, involving self-supervised offline pretraining for trajectory representation and downstream adaptation (online or offline) using the pretrained model based on Transformer architecture. 
For different decision-making environments and tasks, a trajectory can contain different modality information. For example, in robotics, a trajectory usually comprises of proprioceptive states,
camera observations, continuous control actions, task description and goal commands. In game, a trajectory may comprise of image observations, discrete actions and reward scalars.
Formally, a trajectory can be denoted as
\begin{equation}
    \tau = {(x_1^1, x_1^2, ..., x_1^N), ..., (x_T^1, x_T^2, ..., x_T^N)}
\label{eq:traj}
\end{equation}
where $x_t^n$ refers to the n-th modality at the t-th timestep.
Given a sequence of trajectories $\tau$ collected from agent's interactions with environments, the pretraining phase aims to learn a representation function $g:\mathbb{T}\in\mathbb{R}^n\rightarrow\mathbb{Z}\in\mathbb{R}^m$
$(m \ll n)$ to extract useful knowledge from trajectories for downstream adaptation. The knowledge can be temporal information about the same data modality (e.g., $s_t, s_{t+1}$), causal information between different data modalities (e.g., $s_t, a_t$), dynamics information about the environment (e.g., $s_t, a_t, s_{t+1}$), and reward information about the interaction between agents and environments (e.g., $s_t, a_t, r_t$).

During the adaptation phase, the knowledge learned by the pretrained model can help to optimize a learning objective $f_\theta(z)$ in downstream decision-making tasks, such as the value function $V(\pi)$ or $Q(s,a)$, policy function $\pi(a|s)$, dynamics function $T(s,a,s)$, and reward function $r(s,a)$. Compared to learning these objectives from scratch in traditional RL, using the pretrained model will improve the sample efficiency and generalization. The relationship between pretraining and adaptation is shown in Figure \ref{fig:pipeline}.

\section{Data Collection}

\subsection{Pretraining Datasets}
For pretraining, the data can be collected from single or multiple environments and tasks. Recent studies \cite{sun2023smart} show that self-supervised multi-task pretraining can be more beneficial to downstream task learning compared to pretraining with only in-task data, as common knowledge can be extracted from diverse tasks. However, this cannot be always true due to the severe discrepancies between pretraining tasks. For instance, pretraining on multiple Atari games may suffer from performance degradation in downstream game play due to the limited visual feature sharing across the games \cite{stooke2021decoupling}. 

We summarize the sequential decision-making environments and tasks used for self-supervised pretraining data collection in recent studies in Table \ref{tab:env}. To examine the effects of pretraining dataset quality on the self-supervised pretrained model for decision-making, the pretraining datasets can be roughly divided into six categories according to the quality of behavior policy used for data collection. \textbf{Expert}: demonstrations from an expert; \textbf{Near-expert}: expert demonstrations with some action noises; \textbf{Mixed}: transitions collected via multiple checkpoints evenly spread throughout training of an RL algoirthm; \textbf{Weak}: exploratory rollouts of an expert RL algorithm; \textbf{Exploratory}: exploratory rollouts of an exploratory RL algorithm; \textbf{Random}: random interactions with environment. In addition, to test the scalability of Transformer-based pretrained model, the pretraining dataset can be of varying sizes, ranging from several to tens of million transition steps for each pretraining task.

\begin{table*}[t]
\caption{Environments and tasks for self-supervised pretraining for decision-making.}
\label{tab:env}
    \centering
    \begin{tabular}{l|c|c}
    \toprule
    Environment & Task & Continuous/Discrete Action\\
     \hline
    DMControl \cite{tassa2018deepmind}& robotic locomotion and manipulation & continuous\\
    Atari \cite{bellemare2013arcade}  & arcade-style games  & discrete\\
    Gym MuJoCo from D4RL \cite{fu2020d4rl}& simulated locomotion & continuous\\
    Adroit \cite{rajeswaran2017learning} & dexterous manipulation & continuous \\
    Maze2D \cite{fu2020d4rl} & goal-conditioned navigation & continuous\\
    MiniGrid \cite{chevalier2018minimalistic} & 2D map navigation with hierarchical missions & discrete\\
    MiniWorld \cite{MinigridMiniworld23} & 3D visual navigation & continuous\\ 
    Dark Room \cite{zintgraf2019varibad} & 2D goal-oriented navigation & discrete\\
    \bottomrule
    \end{tabular}
\end{table*}

\subsection{Downstream Data}
When fine-tuning the pretrained model on a specific decision-making task, the downstream data can be online interactions with an environment or simply a fixed offline dataset. Different from the large, diverse pretraining dataset where unlabeled and sub-optimal trajectories may compose the most of data, the downstream dataset for fine-tuning is usually in a small size but with reward supervision or expert demonstration for better downstream adaptation. Additionally, the downstream data can be in the same as or different from the pretraining data with respect to the domains and tasks. Generally, unseen domains and tasks during downstream adaptation aim to test the generalization of pretrained representation.

\section{Self-supervised Pretraining}
In this survey, we focus on Transformer-based self-supervised pretraining for decision foundation model. The input to a Transformer architecture is required to be a sequence of discrete tokens, such as word tokens in language or patch tokens in vision. In this section, we first introduce tokenization strategies for learning trajectory tokens and summarize self-supervised pretraining objectives for decision-making proposed in recent work.

\subsection{Tokenization Strategies}
\begin{figure}
    \centering
    \includegraphics[scale=0.35]{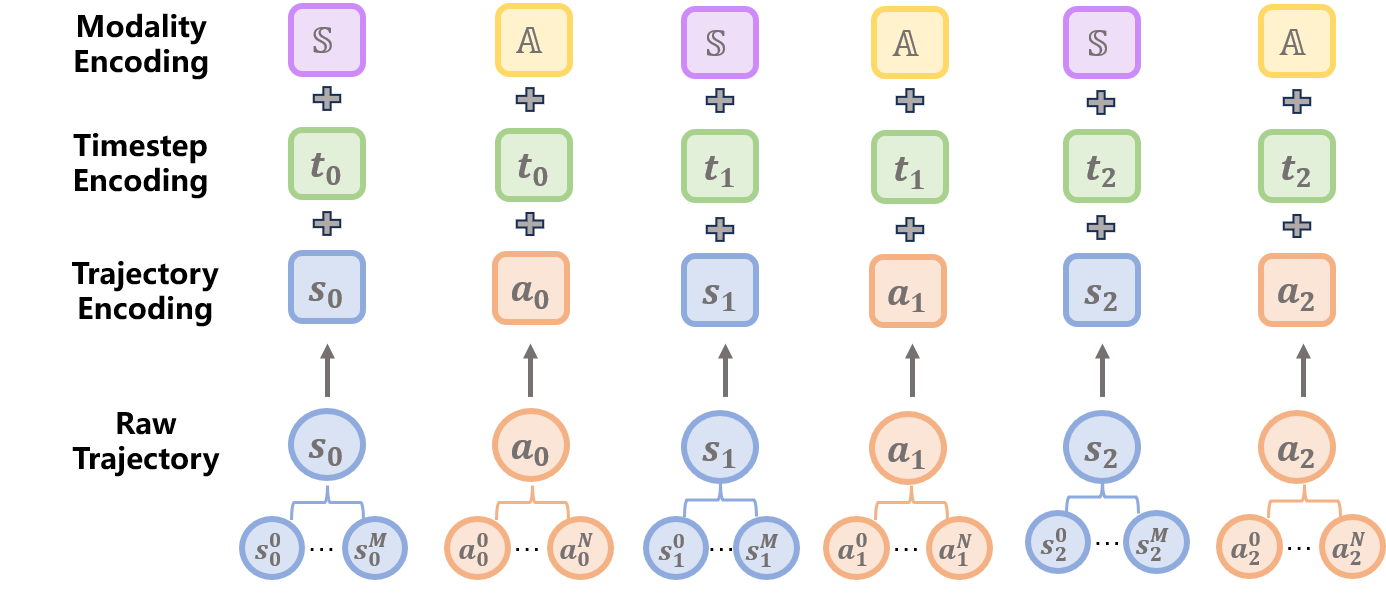}
    \caption{Tokenization strategies for self-supervised pretraining on decision-making tasks. Here the raw trajectory data comprises observations and actions. Note that in-principle, the tokenization can handle any modality. }
    \label{fig:token}
\end{figure}

A trajectory dataset contains multi-modal information as shown in Eq.\ref{eq:traj}. To learn embeddings of different data modalities, separate tokenizer is learned for each modality. In general, tokenization of a trajectory sequence comprises three components:(1) trajectory encoding; (2) timestep encoding; and (3) modality encoding, to allow the Transformer to disambiguate between different elements in the sequence.
\textbf{Trajectory encoding} aims to transform the raw trajectory inputs into a common representation space for tokens. There are two tokenization granularity for trajectory embedding learning: (1) discretization at the level of data modality \cite{sun2023smart,wu2023masked}; and (2) discretization at the level of data dimensions \cite{reed2022generalist,boige2023pasta}. 
 \cite{boige2023pasta} has suggested that discretization at the level of dimensions can improve pretraining performance compared to that at the level of modality, however, the comparative experiments are conducted only on three locomotion tasks. 
\textbf{Timestep encoding} captures absolute or relative positional information. The timestep embedding can be learned via a learned positional embedding layer or a fixed sinusoidal timestep encoder. 
\textbf{Modality encoding} allows the Transformer to distinguish modality types in the trajectory input, which can be learned via a learnable mode-specific encoder \cite{wu2023masked}.
An illustration of tokenization strategies is shown in Figure \ref{fig:token}.
Effects of different tokenization strategies need to be investigated in future work as well as its corresponding factors.

\begin{table*}[t]
    \centering
    \begin{tabular}{l|l|c}
    \toprule
    Objective  & Loss Function  & Base Pattern\\
    \hline
    Next action prediction &$-logP_\theta(a_t|\tau_{0:t-1},s_t)$ & Next token prediction\\
    Reward-conditioned action prediction & $-logP_\theta(a_t|\tau_{0:t-1},s_t, R_t)$&Next token prediction\\
    Future value and reward prediction & $-logP_\theta(a_t, \hat{R_t}, r_t|\tau_{0:t-1},s_t)$&Next token prediction\\
    Future-conditioned action prediction &$-logP_\theta(a_t|\tau_{0:t-1},s_t, z)$ &Next token prediction\\
    Forward dynamics prediction  &$-logP_\theta(s_t|\tau_{0:t-1})$ &Next token prediction\\
    Inverse dynamics prediction  &$-logP_\theta(a_t|s_t, s_{t+1})$ & Next token prediction\\
    Multiple proportion random masking &$-logP_\theta(masked(\tau)|unmasked(\tau)$ &Masked token prediction\\
    Reward-conditioned random mask prediction&
    $-logP_\theta(masked(\tau)|unmasked(\tau), R_0)$ & Masked token prediction\\
    Random masked hindsight prediction  &$-logP_\theta(masked(a)|unmasked(\tau), a_T)$ &Masked token prediction\\
    Random autoregressive mask prediction  &$-logP_\theta(masked(\tau, a_T)|unmasked(\tau))$ &Masked\&Next token prediction\\
    \bottomrule
    \end{tabular}
\caption{Self-supervised pretraining objectives for decision-making based on Transformer architecture.
$P_\theta$ is the prediction of the pretrained model. $\tau_{0:t-1}$ denotes the previous trajectories before timestep $t$. $s_t$ is the state or observation at timestep $t$ and $a_t$ is the action at timestep $t$. $R_t$ is the return (cumulative rewards) at timestep $t$, while $\hat{R_t}$ is the predicted return at timestep $t$ and $r_t$ is the immediate reward at timestep $t$. $g_T$ refers to the goal at the final timestep $T$, and $z$ denotes the encoded future trajectory with the same sequence length as the input trajectory. $s_{t+i}$ is the state or observation at future timestep after $t$ and before the end of sequence length $T$. Note that the conditional variables in each loss function are encoded by either a causal Transformer \protect\cite{brown2020language} when applying next token prediction or a masked Transformer \protect\cite{devlin2018bert} when applying masked token prediction.}
\label{tab:objective}
\end{table*}

\subsection{Pretraining Objectives}
\label{sec:objective}
To leverage large-scale offline dataset with unlabeled samples, self-supervised pretraining has achieved notable success in various domains, such as language \cite{brown2020language}, vision \cite{bai2023sequential}, speech \cite{chen2022wavlm}, and control \cite{yang2021representation}. The self-supervised learning objective is a key component during pretraining, as it allows the model to learn the underlying structure and deep semantics of the data by predicting future information or filling in missing information. Usually, the pretraining objective is designed specifically for certain domain tasks and model architecture. With Transformer which can model long-range dependencies and capture temporal information from input sequence, self-supervised pretraining objectives can be roughly divided into two categories: (1) \textbf{next token prediction}, and (2) \textbf{masked token prediction}. Based on the two patterns, varieties of self-supervised learning objectives are proposed for decision-making task pretraining, as summarized in Table \ref{tab:objective}.

As shown in Table \ref{tab:objective}, the simplest pretraining objective is to predict next action conditioned on previous trajectories and the state or observation at current timestep. It also relies on the least requirements for the data modality of offline pretraining dataset. However, by conditioning on various variables contained in a trajectory sequence, the action prediction can be empowered with different sources of information in the sequential decision process, such as the reward or value signals \cite{chen2021decision,lee2022multi},next state or observation \cite{sun2023smart} and latent future sub-trajectory information \cite{xie2023future}. Besides learning the control information with action as the predicted target, dynamics information can also be exploited during pretraining via state representation learning which has been proven effective in improving sample efficiency in many RL tasks \cite{yang2021representation}. 
In addition to next token prediction, random masking helps to understand the semantics of trajectory data during pretraining by filling in random masked information given context around masked tokens. 
\cite{liu2022masked} randomly mask a proportion of input trajectory data using a randomly sampled mask ratio ranging from 15\% to 95\%.
Besides, varieties of masking schemes have been proposed inspired by different components in RL. Reward-conditioned random mask prediction aims to predicted masked tokens in input trajectory conditioned on unmasked token sequence and the first return token. In  \cite{carroll2022uni}, the first return token can be masked with probability 1/2 and all the subsequent return tokens are masked.
Random masked hindsight prediction \cite{sun2023smart} only learns to recover masked action tokens conditioned on unmasked trajectory tokens and the action at the final timestep $T$ to capture global temporal relations for multi-step control. 
In contrast, random autoregressive mask prediction \cite{wu2023masked} only conditions on unmasked trajectory sequence and learns to recover the masked trajectory tokens with a constraint that the last element in the sequence (e.g., $a_T$) must be necessarily masked to force the pretrained model to be causal at inference time.

In pretraining for decision-making, contrastive self-prediction is another effective self-supervised objective, especially in data augmentation strategies for sample efficiency. Different from sequence modeling, contrastive prediction particularly learns state representation by applying a contrastive loss between encoded state (e.g, $\phi(s_t)$) and target encoded state (e.g, $\phi_{target}(s_{t+i})$). The parameters of target state encoder is non-trainable and usually defined as an exponential moving average of the weights of $\phi$ \cite{stooke2021decoupling,schwarzer2021pretraining,cai2023reprem}.

\section{Downstream Adaptation}
In the pretrain-then-adapt pipeline, a model is first pretrained on a large-scale offline dataset containing trajectories from various domains and decision-making tasks. Then the pretrained model acts as a knowledge base to provide meaningful representation for downstream task learning so as to improve sample efficiency and generalization when adapting to a new task or domain. 
In this section, we first summarize inference tasks in sequential decision-making problems during downstream learning, and introduce two ways of downstream adaptation using a pretrained model for decision-making: (1) fine-tuning and (2) zero-shot generalization.

\subsection{Inference Tasks in Sequential Decision-Making}
\label{sec:infer task}
\textbf{Action prediction}: Predict action based on various trajectory information. For example, reward-conditioned action inference predicts next timestep action using the returns-to-go (RTG) information $R=\sum_{k=t}^{T}r_{k}$.
Goal-conditioned action inference predicts action using the goal information usually provided in goal reaching tasks. It can be a single-goal reaching task given a future state at the end of trajectory sequence $P_\theta(a_t|\tau_{0:t-1},s_t, g_T)$, or a multi-goal reaching task given several goal states at random future timesteps \cite{liu2022masked}. The multi-goal reaching task can also be seen as a variant of waypoint-conditioned action inference\cite{carroll2022uni,badrinath2023waypoint} where some subgoals (or waypoints) are specified at particular timesteps $P_\theta(a_t|\tau_{0:t-1},s_t, s_{t+i})$.

\textbf{Dynamics prediction}: Predict forward or inverse dynamics based on previous or future trajectory information. Forward dynamcis inference aims to predict future state $s_{t+1}$ (and reward $r_{t+1}$) based on current state $s_t$ and action $a_t$ or previous state-action sequences $(s_{0:t}, a_{0:t}$. This inference process models the environment dynamics and it is not restricted to Markovian dynamics in theory.
Inverse dynamics inference aims to predict previous action $a_{t-1}$ (or state $s_{t-1}$) based on current state $s_t$ and previous state $s_{t-1}$ (or action $a_{t-1}$). This inference process tries to recover action (or state) sequences that track desired reference state (or action) trajectories. 

\textbf{Trajectory prediction}: Predict future or past trajectories based on previous or future trajectory information. Particularly, future inference task predicts future state-action sequences based on previous state-action sequences
$P_\theta(\tau_{t:T}|\tau_{0:t-1})$. Conversely, past inference task predict previous state-action sequences based on future state-action sequences $P_\theta(\tau_{0:t-1})|\tau_{t:T})$.

\subsubsection{Downstream Evaluation}
The pretrained model for decision-making is desired to learn representation of various RL components and extract knowledge from the representation to improve sample efficiency and generalization when adapting to a downstream inference task. 
For example, when the inference task is action prediction, a pretrained model can perform policy initialization. When performing dynamics inference, the model can act as a world model for model-based RL algorithms or an inverse dynamics model. For trajectory inference, the pretrained model can provide sufficient trajectory representation to accelerate the learning of traditional RL algorithms.

The performance of decision foundation model can be evaluated from different dimensions depending on the downstream adaptation setting. Generally, the evaluation can be considered from three perspectives: (1) representation learning performance; (2) generalization performance; and (3) robustness performance. For representation learning, the downstream inference tasks and domains are seen during pretraining, while for generalization, the downstream tasks or domains are unseen during pretraining. For robustness, the downstream tasks can be designed to test the model's ability of resilience to corrupted observations or environment variations. \cite{boige2023pasta} evaluates the robust representation by disabling one of eleven sensors or introducing four gravity changes during inference.

\subsection{Fine-Tuning}
Fine-tuning is one way to adapt a pretrained model to a downstream inference task. Particularly, there are two cases demanding fine-tuning: (1) when the pretraining data is mixed comprising a small proportion of near-expert data and a large proportion of exploratory trajectories; and (2) when the pretraining objective is quite different from the learning objective of downstream task. For example, in traditional RL, the learning objective is to maximize the total reward for a task specified by a reward function, whereas the self-supervised pretraining task is usually reward-free.

\subsubsection{Policy Learning Methods}
When fine-tuning a pretrained model to adapt to a new decision-making task, it is required to learn a new policy on the downstream dataset using an RL algorithm, since the pretrained model is to provide useful representation rather than a general policy. When the downstream dataset is in high quality, \textbf{imitation learning} algorithms such as behavior cloning (BC) \cite{sun2023smart,yang2021representation} can be applied whose learning objective is consistent with the action prediction target in pretraining. When the downstream data is an offline dataset, the pretrained model can be fine-tuned with \textbf{offline RL} algorithms, such as RTG-conditioned BC \cite{sun2023smart,xie2023future,lee2022multi,wu2023masked}, BARC \cite{yang2021representation} and TD3-BC \cite{boige2023pasta,liu2022masked}. In online settings, traditional \textbf{online RL} algorithms including SAC \cite{yang2021representation,cai2023reprem}, DQN with forward dynamics losses\cite{schwarzer2021pretraining} and Rainbow \cite{cai2023reprem} can be used to learn the policy for downstream inference task.

\subsubsection{Fine-tuning Methods}
\label{sec:fine-tune method}
When learning a new policy on the downstream dataset, the weights of pretrained model are changed.
According to the proportion of weights changed, fine-tuning methods can be roughly divided into three categories: (1) \textbf{Entirely fine-tuning}: the whole parameters of pretrained model are changed during downstream fine-tuning; (2) \textbf{Probing}: append policy heads to the pretrained model which generate embeddings with frozen weights and only the policy head introduces new parameters. 
\cite{cai2023reprem,schwarzer2021pretraining} show that in some cases freezing pretrained weights performs better than entirely fine-tuning; (3) \textbf{Parameter-efficient fine-tuning (PEFT)}: introduce a limited number of weights to the pretrained model which has seen notable success in fine-tuning large language models, such as LoRA \cite{hu2021lora}, Prefix Tuning \cite{li2021prefix}, P-Tuning\cite{liu-etal-2022-p} and $(IA)^3$ \cite{liu2022few,boige2023pasta}.

\subsection{Zero-shot Generalization}
In addition to fine-tuning, zero-shot generalization directly adapts a pretrained model to downstream tasks without the need to change or introduce any parameters to the pretrained model. This adaptation method performs well when there is a small gap between self-supervised pretraining objectives and downstream inference task objective and the pretraining data is near-expert. Typically, two strategies can be used for zero-shot generalization: (1) \textbf{Aligning} the pretraining objective and downstream inference objective, and (2) \textbf{Prompting} the pretrained model with demonstrations from corresponding downstream task \cite{xu2022prompting,reed2022generalist,laskin2022context}.

For example, using a random masking pretraining objective where the mask ratio varies from goal to goal, \cite{liu2022masked} achieves better performance on goal reaching task compared to using the next token prediction objective. The reason is that the masking pretraining objective naturally fits the goal reaching scenario as the model is required to learn to recover masked actions based on remaining states. 
However, when the downstream inference task is a generation task, the next token prediction objective can be more appropriate for pretraining. By simply prompting the pretrained model with a few state-action pairs of a skill (e.g., walk/stand/run), the model can generate future trajectories in the same skill pattern \cite{liu2022masked}.

\section{Challenges and Future Directions}
Despite the success of pretraining in deep RL \cite{laskin2021urlb,xie2022pretraining,wen2023large}, self-supervised pretraining for decision foundation model is a relatively new area. In this section, we discuss some main challenges and highlight several future directions towards general large-scale self-supervised pretraining for decision foundation model.

\subsection{Pretraining}
Three key components of self-supervised pretraining for decision foundation model include (1) tokenization strategies, (2) pretraining objectives, and (3) data collection.
Currently, the tokenization scheme in pretraining for decision-making mainly borrows ideas from large language or vision models by transforming input data into token sequences. However, one difference between decision-making and language or visual pretraining is that the input trajectory data is multi-modality, whereas the language or visual data is single modality. A challenge then is to align different modalities, which is also a key problem in recent multi-modal pretraining \cite{wang2023large}. In addition, the tokenziation is first proposed for language modeling as language data is naturally discrete token sequences. However, many decision-making tasks handle continuous data (e.g., continuous control tasks for robot learning) and whether existing tokenization methods can fully make use of the information in trajectory data need further exploration.

Besides tokenization, the design of self-supervised pretraining objective also depends on the form of input data as well as the transformer architecture. The next token prediction and masked token prediction have achieved notable success in language pretraining, since the pretraining objective can naturally align the language data and language task well. As a result, all language tasks including translation, classification and question-answering can be unified as a text-to-text transfer learning \cite{raffel2020exploring}.                                    
Inspired by the two pretrainig objectives for language modeling, multiple pretraining objectives have been proposed for decision-making using various RL components as shown in Table \ref{tab:objective}. 
Some studies have shown that using a combination of pretraining objectives can encourage the agent to learn various aspects of environment dynamics \cite{schwarzer2021pretraining} or short-term and long-term control information \cite{sun2023smart}.
However, the relationships between these pretraining objectives are underexplored and whether there exists a general pretraining objective for decision-making tasks remains an open question. 

The size and quality of pretraining data is another important factor for model performance. Currently, most studies on pretraining for decision-making rely on task-relevant expert demonstrations \cite{reed2022generalist,lee2023supervised}, showing that the model scales well with higher-quality pretraining data \cite{schwarzer2021pretraining}.
However, in real-world scenarios, it is more common to collect a large amount of unlabeled and sub-optimal data. Additionally, in traditional online RL setting, the agent is not required to learn from expert demonstrations but to find an optimal policy through interactions with the environment. Therefore, how to learn from large-scale unsupervised sub-optimal data is a key challenge for decision-making pretraining and future work can be explored in reference to traditional RL algorithms such as Q-learning or policy gradient methods.

\subsection{Fine-tuning}
When there is a gap between pretraining objectives and downstream inference objective, fine-tuning the pretrained model can help to adapt to the downstream task. As introduced in section \ref{sec:fine-tune method}, different fine-tuning methods change or introduce parameters in different proportions. Yet the effects of different fine-tuning methods on inference performance have been underestimated. Future work can explore under what conditions that a fine-tuning method should be considered, and the factors influencing the fine-tuning performance, such as the type of inference tasks, the similarities and differences between pretraining objective and fine-tuning objective, or the downstream data structure, etc.

While fine-tuning can help the pretrained model to adapt to a new task, it can also cause catastrophic forgetting of the knowledge acquired from pretraining \cite{french1999catastrophic} as it changes the pretrained weights due to the distribution shift between pretraining data and downstream data. Therefore, the capacity of continual learning \cite{ring1994continual} is also a key challenge for pretraining in decision-making tasks. To resolve this issue, a universal interface between different decision-making tasks can be developed in future work to unify distinct state and action spaces as well as a dynamic model architecture to support continuously added features.

\subsection{Downstream Evaluation}
The performance of a pretrained model can be evaluated by downstream tasks. In language or vision, downstream tasks have been clearly defined and classified as benchmarks for evaluation. For example, BERT \cite{devlin2018bert} has been evaluated on eleven NLP tasks including summarization, question answering and commonsense inference. \cite{bai2023sequential} builds a large vision model and adapts to a range of visual tasks including semantic segmentation, depth estimation, surface normal estimation, edge detection, object detection and colorization. 

However, in pretraining for decision foundation model, there lacks of a unified evaluation framework for fair comparison despite of the URLB benchmark \cite{laskin2021urlb} for online pretraining. On the one hand, downstream inference tasks are usually mixed with policy learning methods and the distinction and relations between them are unclear. On the other hand, evaluation metrics are used for specific decision-making tasks, such as normalized score for games and average return for locomotion tasks. Therefore, a principled evaluation framework for decision foundation model needs to be established in future work.
In addition, to assess the capabilities of pretrained decision models from different dimensions, desired properties of the model need to be defined as well as corresponding downstream tasks.

%% The file named.bst is a bibliograhy style file for BibTeX 0.99c
\bibliographystyle{named}
\bibliography{ijcai24}

\begin{thebibliography}{}

\bibitem[\protect\citeauthoryear{Badrinath \bgroup \em et al.\egroup }{2023}]{badrinath2023waypoint}
Anirudhan Badrinath, Yannis Flet-Berliac, Allen Nie, and Emma Brunskill.
\newblock Waypoint transformer: Reinforcement learning via supervised learning with intermediate targets.
\newblock {\em arXiv preprint arXiv:2306.14069}, 2023.

\bibitem[\protect\citeauthoryear{Bai \bgroup \em et al.\egroup }{2023}]{bai2023sequential}
Yutong Bai, Xinyang Geng, Karttikeya Mangalam, Amir Bar, Alan Yuille, Trevor Darrell, Jitendra Malik, and Alexei~A Efros.
\newblock Sequential modeling enables scalable learning for large vision models.
\newblock {\em arXiv preprint arXiv:2312.00785}, 2023.

\bibitem[\protect\citeauthoryear{Bao \bgroup \em et al.\egroup }{2021}]{bao2021beit}
Hangbo Bao, Li~Dong, Songhao Piao, and Furu Wei.
\newblock Beit: Bert pre-training of image transformers.
\newblock {\em arXiv preprint arXiv:2106.08254}, 2021.

\bibitem[\protect\citeauthoryear{Bellemare \bgroup \em et al.\egroup }{2013}]{bellemare2013arcade}
Marc~G Bellemare, Yavar Naddaf, Joel Veness, and Michael Bowling.
\newblock The arcade learning environment: An evaluation platform for general agents.
\newblock {\em Journal of Artificial Intelligence Research}, 47:253--279, 2013.

\bibitem[\protect\citeauthoryear{Boige \bgroup \em et al.\egroup }{2023}]{boige2023pasta}
Raphael Boige, Yannis Flet-Berliac, Arthur Flajolet, Guillaume Richard, and Thomas Pierrot.
\newblock Pasta: Pretrained action-state transformer agents.
\newblock {\em arXiv preprint arXiv:2307.10936}, 2023.

\bibitem[\protect\citeauthoryear{Brohan \bgroup \em et al.\egroup }{2022}]{brohan2022rt}
Anthony Brohan, Noah Brown, Justice Carbajal, Yevgen Chebotar, Joseph Dabis, Chelsea Finn, Keerthana Gopalakrishnan, Karol Hausman, Alex Herzog, Jasmine Hsu, et~al.
\newblock Rt-1: Robotics transformer for real-world control at scale.
\newblock {\em arXiv preprint arXiv:2212.06817}, 2022.

\bibitem[\protect\citeauthoryear{Brown \bgroup \em et al.\egroup }{2020}]{brown2020language}
Tom Brown, Benjamin Mann, Nick Ryder, Melanie Subbiah, Jared~D Kaplan, Prafulla Dhariwal, Arvind Neelakantan, Pranav Shyam, Girish Sastry, Amanda Askell, et~al.
\newblock Language models are few-shot learners.
\newblock {\em Advances in neural information processing systems}, 33:1877--1901, 2020.

\bibitem[\protect\citeauthoryear{Cai \bgroup \em et al.\egroup }{2023}]{cai2023reprem}
Yuanying Cai, Chuheng Zhang, Wei Shen, Xuyun Zhang, Wenjie Ruan, and Longbo Huang.
\newblock Reprem: Representation pre-training with masked model for reinforcement learning.
\newblock {\em arXiv preprint arXiv:2303.01668}, 2023.

\bibitem[\protect\citeauthoryear{Carroll \bgroup \em et al.\egroup }{2022}]{carroll2022uni}
Micah Carroll, Orr Paradise, Jessy Lin, Raluca Georgescu, Mingfei Sun, David Bignell, Stephanie Milani, Katja Hofmann, Matthew Hausknecht, Anca Dragan, et~al.
\newblock Uni [mask]: Unified inference in sequential decision problems.
\newblock {\em Advances in neural information processing systems}, 35:35365--35378, 2022.

\bibitem[\protect\citeauthoryear{Chandak \bgroup \em et al.\egroup }{2019}]{chandak2019learning}
Yash Chandak, Georgios Theocharous, James Kostas, Scott Jordan, and Philip Thomas.
\newblock Learning action representations for reinforcement learning.
\newblock In {\em International conference on machine learning}, pages 941--950. PMLR, 2019.

\bibitem[\protect\citeauthoryear{Chen \bgroup \em et al.\egroup }{2020}]{chen2020simple}
Ting Chen, Simon Kornblith, Mohammad Norouzi, and Geoffrey Hinton.
\newblock A simple framework for contrastive learning of visual representations.
\newblock In {\em International conference on machine learning}, pages 1597--1607. PMLR, 2020.

\bibitem[\protect\citeauthoryear{Chen \bgroup \em et al.\egroup }{2021}]{chen2021decision}
Lili Chen, Kevin Lu, Aravind Rajeswaran, Kimin Lee, Aditya Grover, Misha Laskin, Pieter Abbeel, Aravind Srinivas, and Igor Mordatch.
\newblock Decision transformer: Reinforcement learning via sequence modeling.
\newblock {\em Advances in neural information processing systems}, 34:15084--15097, 2021.

\bibitem[\protect\citeauthoryear{Chen \bgroup \em et al.\egroup }{2022}]{chen2022wavlm}
Sanyuan Chen, Chengyi Wang, Zhengyang Chen, Yu~Wu, Shujie Liu, Zhuo Chen, Jinyu Li, Naoyuki Kanda, Takuya Yoshioka, Xiong Xiao, et~al.
\newblock Wavlm: Large-scale self-supervised pre-training for full stack speech processing.
\newblock {\em IEEE Journal of Selected Topics in Signal Processing}, 16(6):1505--1518, 2022.

\bibitem[\protect\citeauthoryear{Chevalier-Boisvert \bgroup \em et al.\egroup }{2018}]{chevalier2018minimalistic}
Maxime Chevalier-Boisvert, Lucas Willems, and Suman Pal.
\newblock Minimalistic gridworld environment for openai gym.
\newblock 2018.

\bibitem[\protect\citeauthoryear{Chevalier-Boisvert \bgroup \em et al.\egroup }{2023}]{MinigridMiniworld23}
Maxime Chevalier-Boisvert, Bolun Dai, Mark Towers, Rodrigo de~Lazcano, Lucas Willems, Salem Lahlou, Suman Pal, Pablo~Samuel Castro, and Jordan Terry.
\newblock Minigrid \& miniworld: Modular \& customizable reinforcement learning environments for goal-oriented tasks.
\newblock {\em CoRR}, abs/2306.13831, 2023.

\bibitem[\protect\citeauthoryear{Devlin \bgroup \em et al.\egroup }{2018}]{devlin2018bert}
Jacob Devlin, Ming-Wei Chang, Kenton Lee, and Kristina Toutanova.
\newblock Bert: Pre-training of deep bidirectional transformers for language understanding.
\newblock {\em arXiv preprint arXiv:1810.04805}, 2018.

\bibitem[\protect\citeauthoryear{Eysenbach \bgroup \em et al.\egroup }{2018}]{eysenbach2018diversity}
Benjamin Eysenbach, Abhishek Gupta, Julian Ibarz, and Sergey Levine.
\newblock Diversity is all you need: Learning skills without a reward function.
\newblock {\em arXiv preprint arXiv:1802.06070}, 2018.

\bibitem[\protect\citeauthoryear{French}{1999}]{french1999catastrophic}
Robert~M French.
\newblock Catastrophic forgetting in connectionist networks.
\newblock {\em Trends in cognitive sciences}, 3(4):128--135, 1999.

\bibitem[\protect\citeauthoryear{Fu \bgroup \em et al.\egroup }{2020}]{fu2020d4rl}
Justin Fu, Aviral Kumar, Ofir Nachum, George Tucker, and Sergey Levine.
\newblock D4rl: Datasets for deep data-driven reinforcement learning.
\newblock {\em arXiv preprint arXiv:2004.07219}, 2020.

\bibitem[\protect\citeauthoryear{He \bgroup \em et al.\egroup }{2020}]{he2020momentum}
Kaiming He, Haoqi Fan, Yuxin Wu, Saining Xie, and Ross Girshick.
\newblock Momentum contrast for unsupervised visual representation learning.
\newblock In {\em Proceedings of the IEEE/CVF conference on computer vision and pattern recognition}, pages 9729--9738, 2020.

\bibitem[\protect\citeauthoryear{Hu \bgroup \em et al.\egroup }{2021}]{hu2021lora}
Edward~J Hu, Yelong Shen, Phillip Wallis, Zeyuan Allen-Zhu, Yuanzhi Li, Shean Wang, Lu~Wang, and Weizhu Chen.
\newblock Lora: Low-rank adaptation of large language models.
\newblock {\em arXiv preprint arXiv:2106.09685}, 2021.

\bibitem[\protect\citeauthoryear{Laskin \bgroup \em et al.\egroup }{2021}]{laskin2021urlb}
Michael Laskin, Denis Yarats, Hao Liu, Kimin Lee, Albert Zhan, Kevin Lu, Catherine Cang, Lerrel Pinto, and Pieter Abbeel.
\newblock Urlb: Unsupervised reinforcement learning benchmark.
\newblock {\em arXiv preprint arXiv:2110.15191}, 2021.

\bibitem[\protect\citeauthoryear{Laskin \bgroup \em et al.\egroup }{2022}]{laskin2022context}
Michael Laskin, Luyu Wang, Junhyuk Oh, Emilio Parisotto, Stephen Spencer, Richie Steigerwald, DJ~Strouse, Steven Hansen, Angelos Filos, Ethan Brooks, et~al.
\newblock In-context reinforcement learning with algorithm distillation.
\newblock {\em arXiv preprint arXiv:2210.14215}, 2022.

\bibitem[\protect\citeauthoryear{Lee \bgroup \em et al.\egroup }{2022}]{lee2022multi}
Kuang-Huei Lee, Ofir Nachum, Mengjiao~Sherry Yang, Lisa Lee, Daniel Freeman, Sergio Guadarrama, Ian Fischer, Winnie Xu, Eric Jang, Henryk Michalewski, et~al.
\newblock Multi-game decision transformers.
\newblock {\em Advances in Neural Information Processing Systems}, 35:27921--27936, 2022.

\bibitem[\protect\citeauthoryear{Lee \bgroup \em et al.\egroup }{2023}]{lee2023supervised}
Jonathan~N Lee, Annie Xie, Aldo Pacchiano, Yash Chandak, Chelsea Finn, Ofir Nachum, and Emma Brunskill.
\newblock Supervised pretraining can learn in-context reinforcement learning.
\newblock {\em arXiv preprint arXiv:2306.14892}, 2023.

\bibitem[\protect\citeauthoryear{Levine \bgroup \em et al.\egroup }{2020}]{levine2020offline}
Sergey Levine, Aviral Kumar, George Tucker, and Justin Fu.
\newblock Offline reinforcement learning: Tutorial, review, and perspectives on open problems.
\newblock {\em arXiv preprint arXiv:2005.01643}, 2020.

\bibitem[\protect\citeauthoryear{Li and Liang}{2021}]{li2021prefix}
Xiang~Lisa Li and Percy Liang.
\newblock Prefix-tuning: Optimizing continuous prompts for generation.
\newblock {\em arXiv preprint arXiv:2101.00190}, 2021.

\bibitem[\protect\citeauthoryear{Liang \bgroup \em et al.\egroup }{2019}]{liang2019deep}
Xiaoyuan Liang, Xunsheng Du, Guiling Wang, and Zhu Han.
\newblock A deep reinforcement learning network for traffic light cycle control.
\newblock {\em IEEE Transactions on Vehicular Technology}, 68(2):1243--1253, 2019.

\bibitem[\protect\citeauthoryear{Liu \bgroup \em et al.\egroup }{2022a}]{liu2022masked}
Fangchen Liu, Hao Liu, Aditya Grover, and Pieter Abbeel.
\newblock Masked autoencoding for scalable and generalizable decision making.
\newblock {\em Advances in Neural Information Processing Systems}, 35:12608--12618, 2022.

\bibitem[\protect\citeauthoryear{Liu \bgroup \em et al.\egroup }{2022b}]{liu2022few}
Haokun Liu, Derek Tam, Mohammed Muqeeth, Jay Mohta, Tenghao Huang, Mohit Bansal, and Colin~A Raffel.
\newblock Few-shot parameter-efficient fine-tuning is better and cheaper than in-context learning.
\newblock {\em Advances in Neural Information Processing Systems}, 35:1950--1965, 2022.

\bibitem[\protect\citeauthoryear{Liu \bgroup \em et al.\egroup }{2022c}]{liu-etal-2022-p}
Xiao Liu, Kaixuan Ji, Yicheng Fu, Weng Tam, Zhengxiao Du, Zhilin Yang, and Jie Tang.
\newblock {P}-tuning: Prompt tuning can be comparable to fine-tuning across scales and tasks.
\newblock In Smaranda Muresan, Preslav Nakov, and Aline Villavicencio, editors, {\em Proceedings of the 60th Annual Meeting of the Association for Computational Linguistics (Volume 2: Short Papers)}, pages 61--68, Dublin, Ireland, May 2022. Association for Computational Linguistics.

\bibitem[\protect\citeauthoryear{Ma \bgroup \em et al.\egroup }{2022}]{ma2022vip}
Yecheng~Jason Ma, Shagun Sodhani, Dinesh Jayaraman, Osbert Bastani, Vikash Kumar, and Amy Zhang.
\newblock Vip: Towards universal visual reward and representation via value-implicit pre-training.
\newblock {\em arXiv preprint arXiv:2210.00030}, 2022.

\bibitem[\protect\citeauthoryear{Nakabi and Toivanen}{2021}]{nakabi2021deep}
Taha~Abdelhalim Nakabi and Pekka Toivanen.
\newblock Deep reinforcement learning for energy management in a microgrid with flexible demand.
\newblock {\em Sustainable Energy, Grids and Networks}, 25:100413, 2021.

\bibitem[\protect\citeauthoryear{OpenAI}{2023}]{Achiam2023GPT4TR}
OpenAI.
\newblock Gpt-4 technical report.
\newblock 2023.

\bibitem[\protect\citeauthoryear{Pathak \bgroup \em et al.\egroup }{2017}]{pathak2017curiosity}
Deepak Pathak, Pulkit Agrawal, Alexei~A Efros, and Trevor Darrell.
\newblock Curiosity-driven exploration by self-supervised prediction.
\newblock In {\em International conference on machine learning}, pages 2778--2787. PMLR, 2017.

\bibitem[\protect\citeauthoryear{Raffel \bgroup \em et al.\egroup }{2020}]{raffel2020exploring}
Colin Raffel, Noam Shazeer, Adam Roberts, Katherine Lee, Sharan Narang, Michael Matena, Yanqi Zhou, Wei Li, and Peter~J Liu.
\newblock Exploring the limits of transfer learning with a unified text-to-text transformer.
\newblock {\em The Journal of Machine Learning Research}, 21(1):5485--5551, 2020.

\bibitem[\protect\citeauthoryear{Rajeswaran \bgroup \em et al.\egroup }{2017}]{rajeswaran2017learning}
Aravind Rajeswaran, Vikash Kumar, Abhishek Gupta, Giulia Vezzani, John Schulman, Emanuel Todorov, and Sergey Levine.
\newblock Learning complex dexterous manipulation with deep reinforcement learning and demonstrations.
\newblock {\em arXiv preprint arXiv:1709.10087}, 2017.

\bibitem[\protect\citeauthoryear{Reed \bgroup \em et al.\egroup }{2022}]{reed2022generalist}
Scott Reed, Konrad Zolna, Emilio Parisotto, Sergio~Gomez Colmenarejo, Alexander Novikov, Gabriel Barth-Maron, Mai Gimenez, Yury Sulsky, Jackie Kay, Jost~Tobias Springenberg, et~al.
\newblock A generalist agent.
\newblock {\em arXiv preprint arXiv:2205.06175}, 2022.

\bibitem[\protect\citeauthoryear{Ring and others}{1994}]{ring1994continual}
Mark~Bishop Ring et~al.
\newblock Continual learning in reinforcement environments.
\newblock 1994.

\bibitem[\protect\citeauthoryear{Schwarzer \bgroup \em et al.\egroup }{2021}]{schwarzer2021pretraining}
Max Schwarzer, Nitarshan Rajkumar, Michael Noukhovitch, Ankesh Anand, Laurent Charlin, R~Devon Hjelm, Philip Bachman, and Aaron~C Courville.
\newblock Pretraining representations for data-efficient reinforcement learning.
\newblock {\em Advances in Neural Information Processing Systems}, 34:12686--12699, 2021.

\bibitem[\protect\citeauthoryear{Schwarzer \bgroup \em et al.\egroup }{2023}]{schwarzer2023bigger}
Max Schwarzer, Johan Samir~Obando Ceron, Aaron Courville, Marc~G Bellemare, Rishabh Agarwal, and Pablo~Samuel Castro.
\newblock Bigger, better, faster: Human-level atari with human-level efficiency.
\newblock In {\em International Conference on Machine Learning}, pages 30365--30380. PMLR, 2023.

\bibitem[\protect\citeauthoryear{Silver \bgroup \em et al.\egroup }{2017}]{silver2017mastering}
David Silver, Julian Schrittwieser, Karen Simonyan, Ioannis Antonoglou, Aja Huang, Arthur Guez, Thomas Hubert, Lucas Baker, Matthew Lai, Adrian Bolton, et~al.
\newblock Mastering the game of go without human knowledge.
\newblock {\em nature}, 550(7676):354--359, 2017.

\bibitem[\protect\citeauthoryear{Stooke \bgroup \em et al.\egroup }{2021}]{stooke2021decoupling}
Adam Stooke, Kimin Lee, Pieter Abbeel, and Michael Laskin.
\newblock Decoupling representation learning from reinforcement learning.
\newblock In {\em International Conference on Machine Learning}, pages 9870--9879. PMLR, 2021.

\bibitem[\protect\citeauthoryear{Sun \bgroup \em et al.\egroup }{2023}]{sun2023smart}
Yanchao Sun, Shuang Ma, Ratnesh Madaan, Rogerio Bonatti, Furong Huang, and Ashish Kapoor.
\newblock Smart: Self-supervised multi-task pretraining with control transformers.
\newblock {\em arXiv preprint arXiv:2301.09816}, 2023.

\bibitem[\protect\citeauthoryear{Sutton and Barto}{2018}]{10.5555/3312046}
Richard~S. Sutton and Andrew~G. Barto.
\newblock {\em Reinforcement Learning: An Introduction}.
\newblock A Bradford Book, Cambridge, MA, USA, 2018.

\bibitem[\protect\citeauthoryear{Tang \bgroup \em et al.\egroup }{2022}]{tang2022inputting}
Hongyao Tang, Zhaopeng Meng, Jianye Hao, Chen Chen, Daniel Graves, Dong Li, Changmin Yu, Hangyu Mao, Wulong Liu, Yaodong Yang, et~al.
\newblock What about inputting policy in value function: Policy representation and policy-extended value function approximator.
\newblock In {\em Proceedings of the AAAI Conference on Artificial Intelligence}, volume~36, pages 8441--8449, 2022.

\bibitem[\protect\citeauthoryear{Tassa \bgroup \em et al.\egroup }{2018}]{tassa2018deepmind}
Yuval Tassa, Yotam Doron, Alistair Muldal, Tom Erez, Yazhe Li, Diego de~Las Casas, David Budden, Abbas Abdolmaleki, Josh Merel, Andrew Lefrancq, et~al.
\newblock Deepmind control suite.
\newblock {\em arXiv preprint arXiv:1801.00690}, 2018.

\bibitem[\protect\citeauthoryear{Vaswani \bgroup \em et al.\egroup }{2017}]{vaswani2017attention}
Ashish Vaswani, Noam Shazeer, Niki Parmar, Jakob Uszkoreit, Llion Jones, Aidan~N Gomez, {\L}ukasz Kaiser, and Illia Polosukhin.
\newblock Attention is all you need.
\newblock {\em Advances in neural information processing systems}, 30, 2017.

\bibitem[\protect\citeauthoryear{Wang \bgroup \em et al.\egroup }{2023a}]{wang2023meta}
Mingyang Wang, Zhenshan Bing, Xiangtong Yao, Shuai Wang, Huang Kai, Hang Su, Chenguang Yang, and Alois Knoll.
\newblock Meta-reinforcement learning based on self-supervised task representation learning.
\newblock In {\em Proceedings of the AAAI Conference on Artificial Intelligence}, volume~37, pages 10157--10165, 2023.

\bibitem[\protect\citeauthoryear{Wang \bgroup \em et al.\egroup }{2023b}]{wang2023large}
Xiao Wang, Guangyao Chen, Guangwu Qian, Pengcheng Gao, Xiao-Yong Wei, Yaowei Wang, Yonghong Tian, and Wen Gao.
\newblock Large-scale multi-modal pre-trained models: A comprehensive survey.
\newblock {\em Machine Intelligence Research}, pages 1--36, 2023.

\bibitem[\protect\citeauthoryear{Wen \bgroup \em et al.\egroup }{2023}]{wen2023large}
Muning Wen, Runji Lin, Hanjing Wang, Yaodong Yang, Ying Wen, Luo Mai, Jun Wang, Haifeng Zhang, and Weinan Zhang.
\newblock Large sequence models for sequential decision-making: a survey.
\newblock {\em Frontiers of Computer Science}, 17(6):176349, 2023.

\bibitem[\protect\citeauthoryear{Wu \bgroup \em et al.\egroup }{2023}]{wu2023masked}
Philipp Wu, Arjun Majumdar, Kevin Stone, Yixin Lin, Igor Mordatch, Pieter Abbeel, and Aravind Rajeswaran.
\newblock Masked trajectory models for prediction, representation, and control.
\newblock {\em arXiv preprint arXiv:2305.02968}, 2023.

\bibitem[\protect\citeauthoryear{Xie \bgroup \em et al.\egroup }{2022}]{xie2022pretraining}
Zhihui Xie, Zichuan Lin, Junyou Li, Shuai Li, and Deheng Ye.
\newblock Pretraining in deep reinforcement learning: A survey.
\newblock {\em arXiv preprint arXiv:2211.03959}, 2022.

\bibitem[\protect\citeauthoryear{Xie \bgroup \em et al.\egroup }{2023}]{xie2023future}
Zhihui Xie, Zichuan Lin, Deheng Ye, Qiang Fu, Yang Wei, and Shuai Li.
\newblock Future-conditioned unsupervised pretraining for decision transformer.
\newblock In {\em International Conference on Machine Learning}, pages 38187--38203. PMLR, 2023.

\bibitem[\protect\citeauthoryear{Xu \bgroup \em et al.\egroup }{2022}]{xu2022prompting}
Mengdi Xu, Yikang Shen, Shun Zhang, Yuchen Lu, Ding Zhao, Joshua Tenenbaum, and Chuang Gan.
\newblock Prompting decision transformer for few-shot policy generalization.
\newblock In {\em international conference on machine learning}, pages 24631--24645. PMLR, 2022.

\bibitem[\protect\citeauthoryear{Yang and Nachum}{2021}]{yang2021representation}
Mengjiao Yang and Ofir Nachum.
\newblock Representation matters: Offline pretraining for sequential decision making.
\newblock In {\em International Conference on Machine Learning}, pages 11784--11794. PMLR, 2021.

\bibitem[\protect\citeauthoryear{Yarats \bgroup \em et al.\egroup }{2021}]{yarats2021image}
Denis Yarats, Rob Fergus, and Ilya Kostrikov.
\newblock Image augmentation is all you need: Regularizing deep reinforcement learning from pixels.
\newblock In {\em 9th International Conference on Learning Representations, ICLR 2021}, 2021.

\bibitem[\protect\citeauthoryear{Zhou \bgroup \em et al.\egroup }{2019}]{zhou2019optimization}
Zhenpeng Zhou, Steven Kearnes, Li~Li, Richard~N Zare, and Patrick Riley.
\newblock Optimization of molecules via deep reinforcement learning.
\newblock {\em Scientific reports}, 9(1):10752, 2019.

\bibitem[\protect\citeauthoryear{Zintgraf \bgroup \em et al.\egroup }{2019}]{zintgraf2019varibad}
Luisa Zintgraf, Kyriacos Shiarlis, Maximilian Igl, Sebastian Schulze, Yarin Gal, Katja Hofmann, and Shimon Whiteson.
\newblock Varibad: A very good method for bayes-adaptive deep rl via meta-learning.
\newblock {\em arXiv preprint arXiv:1910.08348}, 2019.

\end{thebibliography}

\end{document}